\documentclass[11pt,a4paper]{article}
\usepackage{times}
\usepackage{latexsym}
\usepackage{caption,comment}
\usepackage{subcaption}
\usepackage{threeparttable}
\usepackage{multirow, array}
\usepackage{booktabs}
\usepackage{amsmath}
\usepackage{graphicx}
\usepackage{textcomp}
\usepackage{url}

\usepackage{natbib}
\usepackage[margin=25mm]{geometry}
\title{Learning to Explicitate Connectives with Seq2Seq Network \\ for Implicit Discourse Relation Classification}

\author{Wei Shi$^\dag$ and Vera Demberg$^{\dag,\ddag}$\\
	$^\dag$Dept. of Language Science and Technology\\ 
	$^\ddag$Dept. of Mathematics and Computer Science, Saarland University\\
    Saarland Informatics Campus, 66123 Saarbr\"ucken, Germany\\
	{\tt\{w.shi, vera\}@coli.uni-saarland.de}}

\date{}

\begin{document}
\maketitle
\thispagestyle{empty}
\pagestyle{empty}

\begin{abstract}
Implicit discourse relation classification is one of the most difficult steps in discourse parsing. The difficulty stems from the fact that the coherence relation must be inferred based on the content of the discourse relational arguments. Therefore, an effective encoding of the relational arguments is of crucial importance.   
We here propose a new model for implicit discourse relation classification, which consists of a classifier, and a sequence-to-sequence model which is trained to generate a representation of the discourse relational arguments by trying to predict the relational arguments including a suitable implicit  connective. Training is possible because such implicit connectives have been annotated as part of the PDTB corpus. Along with a memory network, our model could generate more refined representations for the task. And on the now standard 11-way classification, our method outperforms the previous state of the art systems on the PDTB benchmark on multiple settings including cross validation.  
%   Using cross-validation, we show that further improvements can be obtained by using additional data from explicitated connectives during translation.
\end{abstract}

\section{Introduction}
Discourse relations describe the logical relation between two sentences/clauses. When understanding a text, humans infer discourse relation between text segmentations. They reveal the structural organization of text, and allow for additional inferences. Many natural language processing tasks, such as machine translation, question-answering, automatic summarization, sentiment analysis, and sentence embedding learning, can also profit from having access to discourse relation information. Recent years have seen more and more works on this topic, including two CoNNL shared tasks \citep{xue2015conll,xue2016conll}.

Penn Discourse Tree Bank \citep[PDTB]{Prasad08thepenn} provides lexically-grounded annotations of discourse relations and their two discourse relational arguments (i.e., two text spans). Discourse relations are sometimes signaled by  explicit discourse markers (e.g., \textit{because, but}). Example \ref{eg_1} shows an explicit discourse relation marked by ``because"; the presence of the connective makes it possible to classify the discourse relation with high reliability: \cite{miltsakaki2005experiments} reported an accuracy of 93.09\% for 4-way classification of explicits. 

Discourse relations are however not always marked by an explicit connective. In fact, implicit discourse relations (i.e.~relations not marked by an explicit discourse cue) outnumber explicit discourse relations in naturally occurring text.
Readers can still infer these implicit relations, but automatic classification becomes a lot more difficult in these cases, and represents the main bottleneck in discourse parsing today. Example \ref{eg_2} shows an implicit contrastive relation which can be inferred from the two text spans that have been marked \textit{Arg1} and \textit{Arg2}. When annotating implicit relations in the PDTB, annotators were asked to first insert a connective which expresses the relation, and then annotate the relation label. This procedure was introduced to achieve higher inter-annotator agreement for implicit relations between human annotators. In the approach taken in this paper, our model mimics this procedure by being trained to explicitate the discouse relation, i.e.~to insert a connective as a secondary task.

\begin{enumerate}
	\small
	\item{\label{eg_1}
		\emph{[I refused to pay the cobbler the full \$95]$_{Arg1}$ \underline{\textbf{because}} [He did poor work.]$_{Arg2}$\\
			\vspace{-0.8cm}
			\begin{flushright}
				--- Explicit, Contingency.Cause
			\end{flushright}
		}	
	}
	\vspace{-10pt}
	\item{\label{eg_2}
		\emph{[In the energy mix of the future, bio-energy will also have a key role to play in boosting rural employment and the rural economy in Europe .]$_{Arg1}$ (\textbf{\underline{Implicit = However}}) 
			[At the same time , the promotion of bio-energy must not lead to distortions of competition.]$_{Arg2}$\\
			\vspace{-0.8cm}
			\begin{flushright}
				--- Implicit, Comparison.Contrast
			\end{flushright}
		}
	}

\end{enumerate}
\vspace{-10pt}

The key in implicit discourse relation classification lies in extracting relevant information for the relation label from (the combination of) the discourse relational arguments. Informative signals can consist of surface cues, as well as the semantics of the relational arguments. Statistical approaches have typically relied on linguistically informed features which capture both of these aspects, like temporal markers, polarity tags, Levin verb classes and sentiment lexicons, as well as the Cartesian products of the word tokens in the two arguments \citep{lin2009recognizing}. More recent efforts use distributed representations with neural network architectures \citep{qin2016implicit}. 

The main question in designing neural networks for discourse relation classification is how to get the neural networks to effectively encode the discourse relational arguments such that all of the aspects relevant to the classification of the relation are represented, in particular in the face of very limited amounts of annotated training data, see e.g.~\citet{rutherford2017systematic}. The crucial intuition in the present paper is to make use of the annotated implicit connectives in the PDTB: in addition to the typical relation label classification task, we also train the model to encode and decode the discourse relational arguments, and at the same time predict the implicit connective. This novel secondary task forces the internal representation to more completely encode the semantics of the relational arguments (in order to allow the model to decode later), and to make a more fine-grained classification (predicting the implicit connective) than is necessary for the overall task. This more fine-grained task thus aims to force the model to represent the discourse relational arguments in a way that allows the model to also predict a suitable connective. 
Our overall discourse relation classifier combines representations from the relational arguments as well as the hidden representations generated as part of the encoder-decoder architecture to predict relation labels. What's more, with an explicit memory network, the network also has access to history representations and acquire more explicit context knowledge. We show that our method outperforms previous approaches on the 11-way classification on the PDTB 2.0 benchmark. 

The remaining of the paper is organized as follows: Section 2 discusses related work; Section 3 describes our proposed method; Section 4 gives the training details and experimental results, which is followed by conclusion and future work in section 5.

\vspace{-10pt}
\section{Related Work}
\vspace{-5pt}

\subsection{Implicit Discourse Relation Classification}

Implicit discourse relation recognition is one of the most important components in discourse parsing. With the release of PDTB \citep{Prasad08thepenn}, the largest available corpus which annotates implicit examples with discourse relation labels and implicit connectives, a lot of previous works focused on typical statistical machine learning solutions with manually crafted sparse features \citep{rutherford2014discovering}. 

Recently, neural networks have shown an advantage of dealing with data sparsity problem, and many deep learning methods have been proposed for  discourse parsing, including convolutional \citep{zhang2015shallow}, recurrent \citep{ji2016latent}, character-based \citep{qin2016implicit}, adversarial \citep{qin2017adversarial} neural networks, and pair-aware neural sentence modeling \citep{cai2017pair}. Multi-task learning has also been shown to be beneficial on this task \citep{lan2017multi}.

However, most neural based methods suffer from insufficient annotated data.\cite{wu2016bilingually} extracted bilingual-constrained synthetic implicit data from a sentence-aligned English-Chinese corpus. \cite{shi2017using,shi2018acquiring} proposed to acquire additional training data by exploiting \textit{explicitation} of connectives during translation. Explicitation refers to the fact that translators sometimes add connectives into the text in the target language which were not originally present in the source language. They used explicitated connectives as a source of weak supervision to obtain additional labeled instances, and showed that this extension of the training data leads to substantial performance improvements. 

The huge gap between explicit and implicit relation recognition (namely, 50\% vs. 90\% in 4-way classification) also motivates to incorporate connective information to guide the reasoning process. \citet{zhou2010predicting} used a language model to automatically insert discourse connectives and leverage the information of these predicted connectives. The approach which is most similar in spirit to ours, \citet{qin2017adversarial}, proposed a neural method that incorporates implicit connectives in an adversarial framework to make the representation as similar as connective-augmented one and showed that the inclusion of implicit connectives could help to improve classifier performance.

% tried to enable model to generate connective-augmented discriminative features via an adversarial training procedure. Our work is similar in that we also exploit the annotated implicit connectives, in proposing a sequence-to-sequence architecture which tries to get the model to learn better representations of the relational arguments by training the model to predict arguments as well as the connective.

\vspace{-10pt}
\subsection{Sequence-to-sequence Neural Networks}
Sequence to sequence model is a general end-to-end approach to sequence learning that makes minimal assumptions on the sequence structure, and firstly proposed by \citet{sutskever2014sequence}. It uses multi-layered Long Short-Term Memory (LSTM) or Gated Recurrent Units (GRU) to map the input sequence to a vector with a fixed dimensionality, and then decode the target sequence from the vector with another LSTM / GRU layer. 

Sequence to sequence models allow for flexible input/output dynamics and have enjoyed great success in machine translation and have been broadly used in variety of sequence related tasks such as Question Answering, named entity recognition (NER) / part of speech (POS) tagging and so on. 

If the source and target of a sequence-to-sequence model are exactly the same, it is also called Auto-encoder, \citet{dai2015semi} used a sequence auto-encoder to better represent sentence in an unsupervised way and showed impressive performances on different tasks. The main difference between our model and this one is that we have different input and output (the output contains a connective while the input doesn't). In this way, the model is forced to explicitate implicit relation and try to learn the latent pattern and discourse relation between implicit arguments and connectives and then generate more discriminative representations.

\begin{figure*}
	\centering
	\includegraphics[width=0.7\linewidth, clip=TRUE, trim=0cm 2cm 0cm 1.5cm]{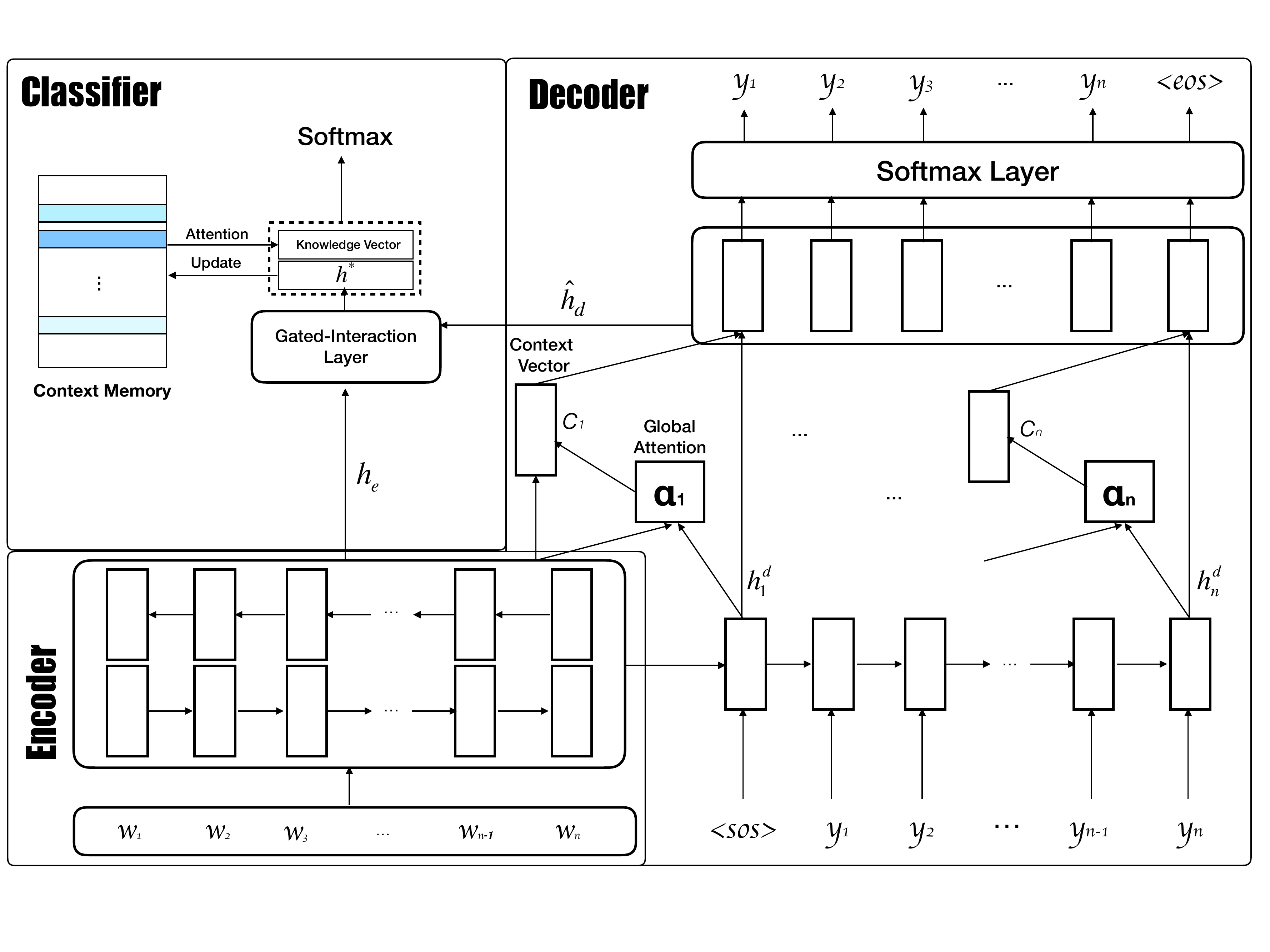}
	\caption{The Architecture of Proposed Model.}
	\label{model}
\end{figure*}

\vspace{-10pt}
\section{Methodology}
\vspace{-5pt}
Our model is based on the sequence-to-sequence model used for machine translation \citep{luong2015effective}, an adaptation of an LSTM \citep{hochreiter1997long} that encodes a variable length input as a fix-length vector, then decodes it into a variable length of outputs. As illustrated in Figure \ref{model}, our model consists of three components: Encoder, Decoder and Discourse Relation Classifier. We here use different LSTMs for the encoding and decoding tasks to help keep the independence between those two parts.

The task of implicit discourse relation recognition is to recognize the senses of the implicit relations, given the two arguments. For each discourse relation instance, The Penn Discourse Tree Bank (PDTB) provides two arguments (\textit{$Arg_1$, $ Arg_2$}) along with the discourse relation (\textit{Rel}) and manually inserted implicit discourse connective (\textit{$Conn_i$}). Here is an implicit example from section 0 in PDTB:

\vspace{-5pt}
\begin{enumerate}
	\footnotesize
	\item[3.]{\label{eg_3}
		\indent $\mathbf{Arg_1}$:~~~This is an old story.\\
		\indent $\mathbf{Arg_2}$:~~~We're talking about years ago before anyone heard of asbestos having any questionable properties.\\
		\indent $\mathbf{Conn_i}$:~in fact\\
		\indent $\mathbf{Rel}$:~~~~~~Expansion.Restatement
	}
\end{enumerate}
\vspace{-5pt}

\noindent During training, the input and target sentences for sequence-to-sequence neural network are $\left[\textit{$Arg_1$}; \textit{$Arg_2$} \right]$ and $\left[\textit{$Arg_1$}; \textit{$Conn_i$}; \textit{$Arg_2$} \right]$ respectively, where ``;'' denotes concatenation.

\vspace{-5pt}
\subsection{Model Architecture}
\subsubsection{Encoder}

Given a sequence of words, an encoder computes a joint representation of the whole sequence.

After mapping tokens to Word2Vec embedding vectors \citep{mikolov2013distributed}, a LSTM recurrent neural network processes a variable-length sequence $x=(x_1, x_2, ..., x_n)$.
%by incrementally adding new contents into a single memory cell, with gates controlling the content to which contents should be memorized, erased or inputed. 
At time step $t$, the state of memory cell $c_t$ and hidden $h_t$ are calculated with the Equations \ref{lstm}:
\begin{equation}\label{lstm}
\small
\begin{gathered}
\left[\begin{array}{c}
i_t \\ f_t \\ o_t \\ \hat{c_t}
\end{array} \right] = \left[\begin{array}{c}
\sigma \\ \sigma \\ \sigma \\ \tanh
\end{array} \right] W \cdot [h_{t-1}, x_t] 
\\
c_t = f_t \odot c_{t-1} + i_t \odot \hat{c_t}
\\
h_t = o_t \odot \tanh(c_t)\\
\end{gathered}
\end{equation}

where $x_t$ is the input at time step $t$, $i$, $f$ and $o$ are the input, forget and output gate activation respectively. $\hat{c_t}$ denotes the current cell state, $\sigma$ is the logistic sigmoid function and $\odot$ denotes element-wise multiplication. The LSTM separates the memory $c$ from the hidden state $h$, which allows for more flexibility incombining new inputs and previous context.

For the sequence modeling tasks, it is beneficial to have access to the past context as well as the future context. Therefore, we chose a bidirectional LSTM as the encoder and the output of the word at time-step $t$ is shown in the Equation~\ref{bi_h}. Here, element-wise sum is used to combine the forward and backward pass outputs.

\begin{equation}\label{bi_h}
\small
h_t = \left[\overrightarrow{h_t} \oplus \overleftarrow{h_t}\right]
\end{equation}

Thus we get the output of encoder:
\begin{equation}
\small
h_{e} = \left[h^e_1, h^e_2, ..., h^e_n\right]
\end{equation}

\vspace{-15pt}
\subsubsection{Decoder}

With the representation from the encoder, the decoder tries to map it back to the targets space and predicts the next words. 

Here we used a separate LSTM recurrent network to predict the target words. During training, target words are fed into the LSTM incrementally and we get the outputs from decoder LSTM:

\begin{equation}
\small
h_{d} = \left[h^d_1, h^d_2, ..., h^d_n\right]
\end{equation}

\vspace{-15pt}
\subsubsection*{Global Attention}
In each time-step in decoding, it's better to consider all the hidden states of the encoder to give the decoder a full view of the source context. So we adopted the global attention mechanism proposed in \citet{luong2015effective}. For time step $t$ in decoding, context vector $c_t$ is the weighted average of $h_{e}$, the weights for each time-step are calculated with $h_t^d$ and $h_{e}$ as illustrated below:
\begin{equation}
\small
\alpha_t =  \frac{\exp({h_t^d}^\top \mathbf{W_{\alpha}} h_{e})}{\sum\limits_{t=1}^n \exp({h_t^d}^\top \mathbf{W_{\alpha}} h_{e})}	
\end{equation}
\begin{equation}
\small
c_t = \alpha h_{e}
\end{equation}

\vspace{-15pt}
\subsubsection*{Word Prediction}
Context vector $c_t$ captured the relevant source side information to help predict the current target word $y_t$. We employ a concatenate layer with activation function $\tanh$ to combine context vector $c_t$ and hidden state of decoder $h^d_t$ at time-step t as follows:
\begin{equation}
\small
\hat{h^d_t} = \tanh(\mathbf{W_c}\left[c_t; h^d_t \right]) 
\end{equation}
Then the predictive vector is fed into the softmax layer to get the predicted distribution $\hat{p}(y_t|s)$ of the current target word.
\begin{equation}
\small
\begin{gathered}   
\hat{p}(y_t|s) = softmax(\mathbf{W}_s \hat{h_d} + \mathbf{b}_s)\\
\hat{y_t} = \arg\max_y\hat{p}(y_t|s)
\end{gathered}
\end{equation}
After decoding, we obtain the predictive vectors for the whole target sequence $\hat{h_d}=\left[h^d_1, h^d_2, ..., h^d_n \right]$. Ideally, it contains the information of exposed implicit connectives.

\vspace{-10pt}
\subsubsection*{Gated Interaction}
In order to predict the coherent discourse relation of the input sequence, we take both the $h_{encoder}$ and the predictive word vectors $h_d$ into account. K-max pooling can ``draw together'' features that are most discriminative and among many positions apart in the sentences, especially on both the two relational arguments in our task here; this method has been proved to be effective in choosing active features in sentence modeling \citep{blunsom2014convolutional}. We employ an average k-max pooling layer which takes average of the top k-max values among the whole time-steps as in Equation~\ref{k_max_1} and \ref{k_max_2}:

\begin{equation}\label{k_max_1}
\small
\bar{h}_e=\frac{1}{k}\sum\limits^{k}_{i=1}topk(h_{e})
\end{equation}
\begin{equation}\label{k_max_2}
\small
\bar{h}_d=\frac{1}{k}\sum\limits^{k}_{i=1}topk(\hat{h^d})
\end{equation}

$\bar{h}_e$ and $\bar{h}_d$ are then combined using a linear layer \citep{lan2017multi}. As illustrated in Equation \ref{equ_interaction}, the linear layer acts as a gate to determine how much information from the sequence-to-sequence network should be mixed into the original sentence's representations from the encoder. Compared with bilinear layer, it also has less parameters and allows us to use high dimensional word vectors.

\begin{equation}
\small
\label{equ_interaction}
h^* = \bar{h}_e \oplus \sigma(\mathbf{W}_i \bar{h}_d + \mathbf{b}_i)
\end{equation}

\vspace{-15pt}
\subsubsection*{Explicit Context Knowledge}
To further capture common knowledge in contexts, we here employ a memory network proposed in \citet{liu2018learning}, to get explicit context representations of contexts training examples. We use a memory matrix $M \in R^{K \times N}$, where $K, N$ denote hidden size and number of training instances respectively. During training, the memory matrix remembers the information of training examples and then retrieves them when predicting labels.

Given a representation $h^*$ from the interaction layer, we generate a \textbf{knowledge vector} by weighted memory reading:
\begin{equation}
\small
	k = M softmax(M^Th^*)
\end{equation}

We here use dot product attention, which is faster and space-efficient than additive attention, to calculate the scores for each training instances. The scores are normalized with a softmax layer and the final knowledge vector is a weighted sum of the columns in memory matrix $M$.

Afterwards, the model predicts the discourse relation using a softmax layer.

\begin{equation}
\small
\begin{gathered}   
\hat{p}(r|s) = softmax(\mathbf{W}_r [k; h^*] + \mathbf{b}_r)\\
\hat{r} = \arg\max_y\hat{p}(r|s)
\end{gathered}
\end{equation}

\vspace{-20pt}
\subsection{Multi-objectives}

In our model, the decoder and the discourse relation classifier have different objectives. For the decoder, the objective consists of predicting the target word at each time-step. The loss function is calculated with masked cross entropy with $\mathtt{L2}$ regularization, as follows:

\begin{equation}
\small
\mathit{Loss_{de}} = -\frac{1}{n}\sum\limits^n_{t=1}y_t\log(\hat{p_y}) + \frac{\lambda}{2}\parallel\theta_{de}\parallel^2_2
\label{decoder_loss}
\end{equation}
where $y_t$ is one-hot represented ground truth of target words, $\hat{p_y}$ is the estimated probabilities for each words in vocabulary by softmax layer, $n$ denotes the length of target sentence. $\lambda$ is a hyper-parameter of $L2$ regularization and $\theta$ is the parameter set.

The objective of the discourse relation classifier consists of predicting the discourse relations. A reasonable training objective for multiple classes is the categorical cross-entropy loss. The loss is formulated as:
\begin{equation}
\small
\mathit{Loss_{cl}} = -\frac{1}{m}\sum\limits^m_{i=1}r_i\log(\hat{p_r}) + \frac{\lambda}{2}\parallel\theta_{cl}\parallel^2_2
\end{equation}
where $r_i$ is one-hot represented ground truth of discourse relation labels, $\hat{p_r}$ denotes the predicted probabilities for each relation class by softmax layer, $m$ is the number of target classes. Just like above, $\lambda$ is a hyper-parameter of $L2$ regularization.

For the overall loss of the whole model, we set another hyper-parameter $w$ to give these two objective functions different weights. Larger $w$ means that more importance is placed on the decoder task.
\begin{equation}
\small
\mathit{Loss} = \mathit{w} \cdot \mathit{Loss_{de}} + \mathit{(1-w)} \cdot \mathit{Loss_{cl}}
\label{weighted_losses}
\end{equation}

\vspace{-20pt}
\subsection{Model Training}
To train our model, the training objective is defined by the loss function we introduced above. We use Adaptive Moment Estimation (Adam) \citep{kingma2014adam} with different learning rate for different parts of the model as our optimizer. Dropout layers are applied after the embedding layer and also on the top feature vector before the softmax layer in the classifier. We also employ $L_2$ regularization with small $\lambda$ in our objective functions for preventing over-fitting. The values of the hyper-parameters, are provided in Table \ref{table:hyper}. The model is trained firstly to minimize the loss in Equation \ref{decoder_loss} until convergence, we use scheduled sampling \citep{bengio2015scheduled} during training to avoid  ``teacher-forcing problem". And then to minimize the joint loss in Equation \ref{weighted_losses} to train the implicit discourse relation classifier.

\vspace{-10pt}
\section{Experiments and Results}
\vspace{-5pt}

\subsection{Experimental Setup} \label{exp_setup}
We evaluate our model on the PDTB. While early work only evaluated classification performance between the four main PDTB relation classes, more recent work including the CoNLL 2015 and 2016 shared tasks on Shallow Discourse Parsing \citep{xue2015conll,xue2016conll} have set the standard to second-level classification. The second-level classification is more useful for most downstream tasks. Following other works we directly compare to in our evaluation, we here use the setting where AltLex, EntRel and NoRel tags are ignored.
About 2.2\% of the implicit relation instances in PDTB have been annotated with two relations, these are considered as two training instances. 
% We remove 5 relations with too few instances (\textsc{Condition, Pragmatic condition, Pragmatic contrast, Pragmatic concession} and \textsc{Exception}) following the setting used in \cite{lin2009recognizing}; this procedure then yields the standard set of 11 level-2 classes. 
% In order to allow for best possible comparison with related work, we here evaluate both on 11-way classification as well as the 4-way classification.

%For the four-way distinction, we follow the setting in previous works to split the data into training set (section 2-20), development set (section 0-1) and test set (section 21-22), while for second-level task, 
To allow for full comparability to earlier work, we here report results for three different settings.
The first one is denoted as PDTB-Lin \citep{lin2009recognizing}; it uses sections 2-21 for training, 22 as dev and section 23 as test set. The second one is labeled PDTB-Ji \citep{ji2015one}, and uses sections 2-20 for training, 0-1 as dev and evaluates on sections 21-22. Our third setting follows the recommendations of \citet{shi2017need}, and performs 10-fold cross validation on the whole corpus (sections 0-23).
Table \ref{table:train_num} shows the number of instances in train, development and test set in different settings.

\begin{table}[t]
	\centering
	\scriptsize
	\begin{tabular}{|c|ccc|}
		\hline
		Settings & Train & Dev & Test \\ \hline
		PDTB-Lin & 13351 & 515 & 766 \\ 
		PDTB-Ji  & 12826 & 1165 & 1039 \\ 
		Cross valid. per fold avg.& 12085 & 1486	& 1486\footnotemark  \\ \hline
	\end{tabular}
	\caption{Numbers of train, development and test set on different settings for 11-way classification task. Instances annotated with two labels are double-counted and some relations with few instances have been removed.}
	\label{table:train_num}
	\vspace{-10pt}
\end{table}
\footnotetext{Cross-validation allows us to test on all 15057 instances.}

The advantage of the cross validation approach is that it addresses problems related to the small corpus size, as it reports model performance across all folds. This is important, because the most frequently used test set (PDTB-Lin) contains less than 800 instances; taken together with a lack in the community to report mean and standard deviations from multiple runs of neural networks \citep{reimers2018comparing}, the small size of the test set makes reported results potentially unreliable.

\vspace{-10pt}
\subsubsection*{Preprocessing}
\vspace{-5pt}
We first convert tokens in PDTB to lowercase and normalize strings, which removes special characters. The word embeddings used for initializing the word representations are trained with the CBOW architecture in \textit{Word2Vec}\footnote{\url{https://code.google.com/archive/p/word2vec/}} \citep{mikolov2013distributed} on PDTB training set. All the weights in the model are initialized with uniform random.

To better locate the connective positions in the target side, we use two position indicators ($\langle conn \rangle$, $\langle /conn \rangle$) which specify the starting and ending of the connectives \citep{zhou2016attention}, which also indicate the spans of discourse arguments.

Since our main task here is not generating arguments, it is better to have representations generated by correct words rather than by wrongly predicted ones. So at test time, instead of using the predicted word from previous time step as current input, we use the source sentence as the decoder's input and target. As the implicit connective is not available at test time, we  use a random vector, which we used as ``impl\_conn'' in Figure~\ref{fig:att_weights}, as a placeholder to inform the sequence that the upcoming word should be a connective.

\vspace{-10pt}
\subsubsection*{Hyper-parameters}
\vspace{-5pt}

There are several hyper-parameters in our model, including dimension of word vectors $d$, two dropout rates after embedding layer $q_1$ and before softmax layer $q_2$, two learning rates for encoder-decoder $lr_1$ and for classifier $lr_2$, top $k$ for k-max pooling layer, different weights $w$ for losses in Equation \eqref{weighted_losses} and $\lambda$ denotes the coefficient of regularizer, which controls the importance of the regularization term, as shown in Table \ref{table:hyper}.

\begin{table}[!b]
	\centering
	\scriptsize
	\begin{tabular}{|c|c|c|c|c|c|c|c|}
		\hline
		$d$ & $q_1$ & $q_2$ & ${lr}_1$    & ${lr}_2$  & $k$ & $w$ & $\lambda$\\ \hline 
		100 & 0.5   & 0.2   & $2.5e^{-3}$ & $5e^{-3}$ &  5  & 0.2 & $5e^{-4}$ \\ \hline
	\end{tabular}
	\caption{Hyper-parameter settings.}
	\label{table:hyper}
\end{table}

%\vspace{-10pt}
\subsection{Experimental Results}

% \subsubsection{Second-level Multi-class Classification}

We compare our models with six previous methods, as shown in Table \ref{table:performance_level2}. The baselines contain feature-based methods \citep{lin2009recognizing}, state-of-the-art neural networks \citep{qin2016implicit,cai2017pair}, including the adversarial neural network that also exploits the annotated implicit connectives \citep{qin2017adversarial}, as well as the data extension method based on using explicitated connectives from translation to other languages \citep{shi2017using}.

Additionally, we ablate our model by taking out the prediction of the implicit connective in the sequence to sequence model. The resulting model is labeled Auto-Encoder in Table \ref{table:performance_level2}. And seq2seq network without knowledge memory, which means we use the output of gated interaction layer to predict the label directly, as denoted as Seq2Seq w/o Mem Net.

\begin{table*}
	\small
	\centering
	\begin{threeparttable}
		\begin{tabular}{llll}
			\toprule
			Methods & PDTB-Lin & PDTB-Ji & Cross Validation \\
			\midrule
			Majority class & 26.11 & 26.18 & 25.59 \\
			\citet{lin2009recognizing} & 40.20 & - & -\\
			%			\citet{lin2009recognizing} + Brown Cluster & - & 40.66 & -\\
			%			\citet{ji2015one} & - & 44.59 & - \\
			\citet{qin2016implicit} & 43.81 & 45.04 & - \\
			\citet{cai2017pair} & - & 45.81 & - \\
			\citet{qin2017adversarial} & 44.65 & \textbf{46.23} & - \\
			\citet{shi2017using} (with extra data) & \textbf{45.50} & - & \textbf{37.84} \\
			%     \hspace{2.5cm} (without extra data) & 34.32 & - & 30.01\\
			\midrule
			Encoder only (Bi-LSTM) \citep{shi2017using} & 34.32 & - & 30.01 \\
			Auto-Encoder	& 43.86		& 45.43	  & 39.50						\\
			Seq2Seq w/o Mem Net & 45.75 & 47.05 & 40.29\\
			Proposed Method & \textbf{45.82} & \textbf{47.83} & \textbf{41.29} \\
			%			Proposed method (with extra data) & 43.73 & 44.85 & 43.59 \\
			
			\bottomrule
		\end{tabular}
		%     \begin{tablenotes}
		%     	\item[1] ``-'' means no result currently.
		%     \end{tablenotes}
		\captionsetup{width=15cm}
		\caption{Accuracy (\%) of implicit discourse relations on PDTB-Lin, PDTB-Ji and Cross Validation Settings for multi-class classification.}
		\label{table:performance_level2}
	\end{threeparttable}
\end{table*}

Our proposed model outperforms the other models in each of the settings. Compared with performances in \citet{qin2017adversarial}, although we share the similar idea of extracting highly discriminative features by generating connective-augmented representations for implicit discourse relations, our method improves about 1.2\% on setting PDTB-Lin and 1.6\% on the PDTB-Ji setting. 
The importance of the implicit connective is also illustrated by the fact that the ``Auto-Encoder'' model, which is identical to our model except it does not predict the implicit connective, performs worse than the model which does. This confirms our initial hypothesis that training with implicit connectives helps to expose the latent discriminative features in the relational arguments, and generates more refined semantic representation. It also means that, to some extent, purely increasing the size of tunable parameters is not always helpful in this task and trying to predict implicit connectives in the decoder does indeed help the model extract more discriminative features for this task. What's more, we can also see that without the memory network, the performances are also worse, it shows that with the concatenation of knowledge vector, the training instance may be capable of finding related instances to get common knowledge for predicting implicit relations. As \citet{shi2017need} argued that it is risky to conclude with testing on such small test set, we also run cross-validation on the whole PDTB. From Table \ref{table:performance_level2}, we have the same conclusion with the effectiveness of our method, which outperformed the baseline (Bi-LSTM) with more than 11\% points and 3\% compared with \citet{shi2017using} even though they have used a very large extra corpus.

\begin{figure*}[!t]
	\small
	\centering
	\begin{subfigure}{.33\textwidth}
		\centering
		\includegraphics[width=\linewidth]{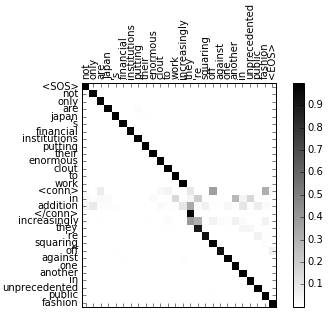}
	\end{subfigure}%
	\begin{subfigure}{.33\textwidth}
		\centering
		\includegraphics[width=\linewidth]{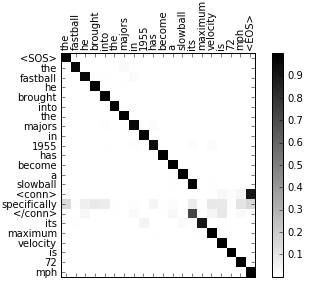}
	\end{subfigure}%
	% \begin{subfigure}{.33\textwidth}
	% 	\centering
	% 	\includegraphics[width=\linewidth]{test_1.png}
	% \end{subfigure}%
	\begin{subfigure}{.33\textwidth}
		\centering
		\includegraphics[width=\linewidth]{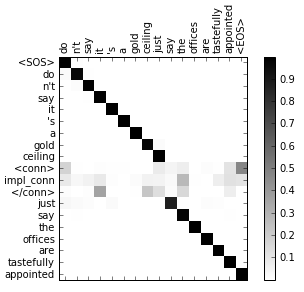}
	\end{subfigure}
	\caption{Visualization of attention weights during predicting target sentence in train and test, x-axis denotes the source sentence and the y-axis is the targets. First two figures are examples from training set with implicit connectives inside, while the following one, in which the implicit connective has been replaced by the word filler ``impl\_conn", is from test.}
	\label{fig:att_weights}
\end{figure*}

For the sake of obtaining a better intuition on how the global attention works in our model, Figure \ref{fig:att_weights} demonstrates the weights of different time-steps in attention layer from the decoder. The weights show how much importance the word attached to the source words while predicting target words. We can see that without the connective in the target side of test, the word filler still works as a connective to help predict the upcoming words. For instance, the true discourse relation for the right-hand example is \textit{Expansion.Alternative}, at the word filler's time-step, it attached more importance on the negation ``don't'' and ``tastefully appointed". It means the current representation could grasp the key information and try to focus on the important words to help with the task. Here we see plenty room for adapting this model to discourse connective prediction task, we would like to leave this to the future work. 

We also try to figure out which instances' representations have been chosen from the memory matrix while predicting. Table \ref{table:mem_att_examples} shows two examples and their context instances with top 2 memory attentions among the whole training set. We can see that both examples show that the memory attention attached more importance on the same relations. This means that with the Context Memory, the model could facilitate the discourse relation prediction by choosing examples that share similar semantic representation and discourse relation during prediction.

\begin{table}[!t]
	\footnotesize
	\centering
	\begin{tabular}{|p{15.5cm}|}
		\hline \vspace{1pt}
		\textit{In recent years, U.S. steelmakers have supplied about 80\% of the 100 million tons of steel used annually by the nation. (\textbf{in addition,}) Of the remaining 20\% needed, the steel-quota negotiations allocate about 15\% to foreign suppliers.}		\\
		\hfill --- Expansion.Conjunction \\ \\ \vspace{-0.5cm}
		
		1. The average debt of medical school graduates who borrowed to pay for their education jumped 10\% to \$42,374 this year from \$38,489 in 1988, says the Association of American Medical Colleges. (\textbf{furthermore}) that's 115\% more than in 1981  \\
		\hfill --- Expansion.Conjunction \\

		2. ... he rigged up an alarm system, including a portable beeper, to alert him when Sventek came on the line. (\textbf{and}) Some nights he slept under his desk.	\\
		\hfill --- Expansion.Conjunction \\ 
        
       	\hline \vspace{1pt}
		\textit{Prices for capital equipment rose a hefty 1.1\% in September, while prices for home electronic equipment fell 1.1\%. (\textbf{Meanwhile,}) food prices declined 0.6\%, after climbing 0.3\% in August.} \\
		\hfill --- Comparison.Contrast     \\ \\   \vspace{-0.5cm}
		
        1. Lloyd's overblown bureaucracy also hampers efforts to update marketing strategies. (\textbf{Although}) some underwriters have been pressing for years to tap the low-margin business by selling some policies directly to consumers. \\
		\hfill --- Comparison.Contrast
        
		2. Valley National "isn't out of the woods yet. (\textbf{Specifically}), the key will be whether Arizona real estate turns around or at least stabilizes \\
		\hfill --- Expansion.Restatement
		\\ \hline
	\end{tabular}
	\caption{Example of attention in Context Knowledge Memory. The sentences in italic are from PDTB test set and following 2 instances are the ones with top 2 attention weights from training set.}
	\label{table:mem_att_examples}
\end{table}

\begin{table}[!t]
\small
\centering
	\begin{tabular}{|c|c|c|c|}
    \hline
	Relation     & Train &  Dev &  Test \\ \hline
    Comparison 	 & 1855  &  189 &  145 	\\ 
    Contingency  & 3235  &  281 &  273 	\\ 
    Expansion	 & 6673	 &  638	&  538  \\ 
    Temporal	 & 582   &  48  &  55	\\ \hline
    Total & 12345& 1156 & 1011\\ \hline
	\end{tabular}
\caption{Distribution of top-level implicit discourse relations in the PDTB.}
\label{table:relation_num_top}
\end{table}

\begin{table*}[!t]
 	\centering
 	\scriptsize
 	\begin{tabular}{|l|c|c|c|c|c|c|}
 		\hline
 		\multirow{2}{*}{Methods} 
 		& \multicolumn{2}{c|}{Four-ways} & \multicolumn{4}{c|}{One-Versus-all Binary ($F_1$)} \\ \cline{2-7}
 		& $F_1$ & Acc. & Comp. & Cont. & Expa. & Temp. \\ \hline
 		\citet{rutherford2014discovering} & 38.40 & 55.50 & 39.70 & 54.42 & 70.23 & 28.69 \\
 		\citet{qin2016stacking} & - & - & \textbf{41.55} & 57.32 & 71.50 & 35.43 \\
 		\citet{liu2016implicit} & 44.98 & 57.27 & 37.91 & 55.88 & 69.97 & 37.17 \\
 		\citet{ji2016latent} & 42.30 & \textbf{59.50} & - & - & - & - \\
 		\citet{liu2016recognizing} & 46.29 & 57.17 & 36.70 & 54.48 & 70.43 & \textbf{38.84} \\
 		\citet{qin2017adversarial} & - & - & 40.87 & 54.46 & 72.38 & 36.20 \\
 		\citet{lan2017multi} & \textbf{47.80} & 57.39 & 40.73 & \textbf{58.96} & \textbf{72.47} & 38.50 \\ \hline
 		Our method & 46.40 & \textbf{61.42} & \textbf{41.83} & \textbf{62.07} & 69.58 & 35.72 \\ \hline
 	\end{tabular}
 	\captionsetup{width=\linewidth}
 	\caption{Comparison of $F_1$ scores (\%) and Accuracy (\%) with the State-of-the-art Approaches for four-ways and one-versus-all binary classification on PDTB. Comp., Cont., Expa. and Temp. stand for Comparison, Contingency, Expansion and Temporal respectively.}
 	\label{performance_level1}
 \end{table*}

\subsubsection{Top-level Binary and 4-way Classification}

A lot of the recent works in PDTB relation recognition have focused on first level relations, both on binary and 4-ways classification. We also report the performance on level-one relation classification for more comparison to prior works. As described above, we followed the conventional experimental settings \citep{rutherford2015improving,liu2016recognizing} as closely as possible. Table \ref{table:relation_num_top} shows the distribution of top-level implicit discourse relation in PDTB, it's worth noticing that there are only 55 instances for Temporal Relation in the test set.

To make the results comparable with previous work, we report the $F_1$ score for four binary classifications and both $F_1$ and Accuracy for 4-way classification, which can be found in Table \ref{performance_level1}. We can see that our method outperforms all alternatives on \textsc{Comparison} and \textsc{Contingency}, and obtain comparable scores with the state-of-the-art in others. For 4-way classification, we got the best accuracy and second-best $F_1$ with around 2\% better than in \citet{ji2016latent}.

\vspace{-10pt}
\section{Conclusion and Future Work}

We present in this paper a novel neural method trying to integrate implicit connectives into the representation of implicit discourse relations with a joint learning framework of sequence-to-sequence network. 
We conduct experiments with different settings on PDTB benchmark, the results show that our proposed method can achieve state-of-the-art performance on recognizing the implicit discourse relations and the improvements are not only brought by the increasing number of parameters. The model also has great potential abilities in implicit connective prediction in the future. 

Our proposed method shares similar spirit with previous work in \citet{zhou2010predicting}, who also tried to leverage implicit connectives to help extract discriminative features from implicit discourse instances. Comparing with the adversarial method proposed by \citet{qin2017adversarial}, our proposed model more closely mimics humans' annotation process of implicit discourse relations and is trained to directly explicitate the implicit relations before classification. With the representation of the original implicit sentence and the explicitated one from decoder, and the help of the explicit knowledge vector from memory network, the implicit relation could be classified with higher accuracy. 

Although our method has not been trained as a generative model in our experiments, we can see potential for applying it to generative tasks. With more annotated data, minor modification and fine-tuned training, we believe our proposed method could also be applied to tasks like implicit discourse connective prediction, or argument generation in the future.

\vspace{-10pt}
\section{Acknowledgments}
This work was supported by German Research Foundation (DFG) as part of SFB 1102 ``Information Density and Linguistic Encoding''. We would like to thank the anonymous reviewers for their careful reading and insightful comments.

\bibliography{iwcs2019}
\bibliographystyle{chicago}

\appendix

% \section{Supplemental Material}
\end{document}